\pdfoutput=1

\documentclass[11pt]{article}

\usepackage{ACL2023}

\usepackage{times}
\usepackage{latexsym}

\usepackage[T1]{fontenc}

\usepackage[utf8]{inputenc}

\usepackage{microtype}

\usepackage{inconsolata}

\usepackage{booktabs}
\usepackage{amsfonts} 
\usepackage{amsmath}
\usepackage{bm}
\usepackage{caption}
\usepackage{colortbl}
\usepackage{tabularx}
\usepackage{multirow}
\usepackage{multicol}
\usepackage{graphicx}
\usepackage{blindtext}
\usepackage{hyperref}
\usepackage{xcolor, colortbl}
\usepackage{relsize}
\usepackage{CJKutf8}
\usepackage{pdfpages}
\usepackage{subcaption}

\usepackage{amssymb}
\usepackage{listings}
\usepackage{tcolorbox}
\usepackage{cleveref}
\usepackage{fancyhdr}

\title{\hspace{-0.4cm}EmoBench: Evaluating the Emotional Intelligence of Large Language Models}

\author{Sahand Sabour\textsuperscript{\rm 1} \quad Siyang Liu\textsuperscript{\rm 2} \quad Zheyuan Zhang\textsuperscript{\rm 3}\quad June M. Liu\textsuperscript{\rm 4}\quad Jinfeng Zhou\textsuperscript{\rm 1}\\ \vspace{0.1cm} {\bf Alvionna S. Sunaryo\textsuperscript{\rm 1}\quad Juanzi Li\textsuperscript{\rm 3}\quad Tatia M.C. Lee\textsuperscript{\rm 4} \quad Rada Mihalcea\textsuperscript{\rm 2}  \quad Minlie Huang\textsuperscript{\rm 1}}
\\ \\
\textsuperscript{\rm 1} The CoAI Group, DCST, Institute for Artificial Intelligence, Tsinghua University, Beijing, China \\
\textsuperscript{\rm 2} The LIT Group, Department of Computer Science and Engineering, University of Michigan, Ann Arbor \\
\textsuperscript{\rm 3} The Knowledge Engineering Group (KEG), DCST, Tsinghua University, Beijing, China \\
  \textsuperscript{\rm 4} The Laboratory of Neuropsychology and Human
Neuroscience, HKU, Hong Kong SAR, China \\
  \texttt{sahandfer@gmail.com}, \texttt{aihuang@tsinghua.edu.cn}}

\begin{document}

\maketitle

\begin{abstract}
Recent advances in Large Language Models (LLMs) have highlighted the need for robust, comprehensive, and challenging benchmarks.
Yet, research on evaluating their Emotional Intelligence (EI) is considerably limited.
Existing benchmarks have two major shortcomings: first, they mainly focus on emotion recognition, neglecting essential EI capabilities such as emotion regulation and thought facilitation through emotion understanding; second, they are primarily constructed from existing datasets, which include frequent patterns, explicit information, and annotation errors, leading to unreliable evaluation.
We propose \textsc{EmoBench}, a benchmark that draws upon established psychological theories and proposes a comprehensive definition for machine EI, including Emotional Understanding and Emotional Application.
\textsc{EmoBench} includes a set of 400 hand-crafted questions in English and Chinese, which are meticulously designed to require thorough reasoning and understanding.
Our findings reveal a considerable gap between the EI of existing LLMs and the average human, highlighting a promising direction for future research.
Our code and data are publicly available at \url{https://github.com/Sahandfer/EmoBench}.
\end{abstract}

\section{Introduction}
\begin{figure}[!ht]
    \centering
    \includegraphics[width=\linewidth]{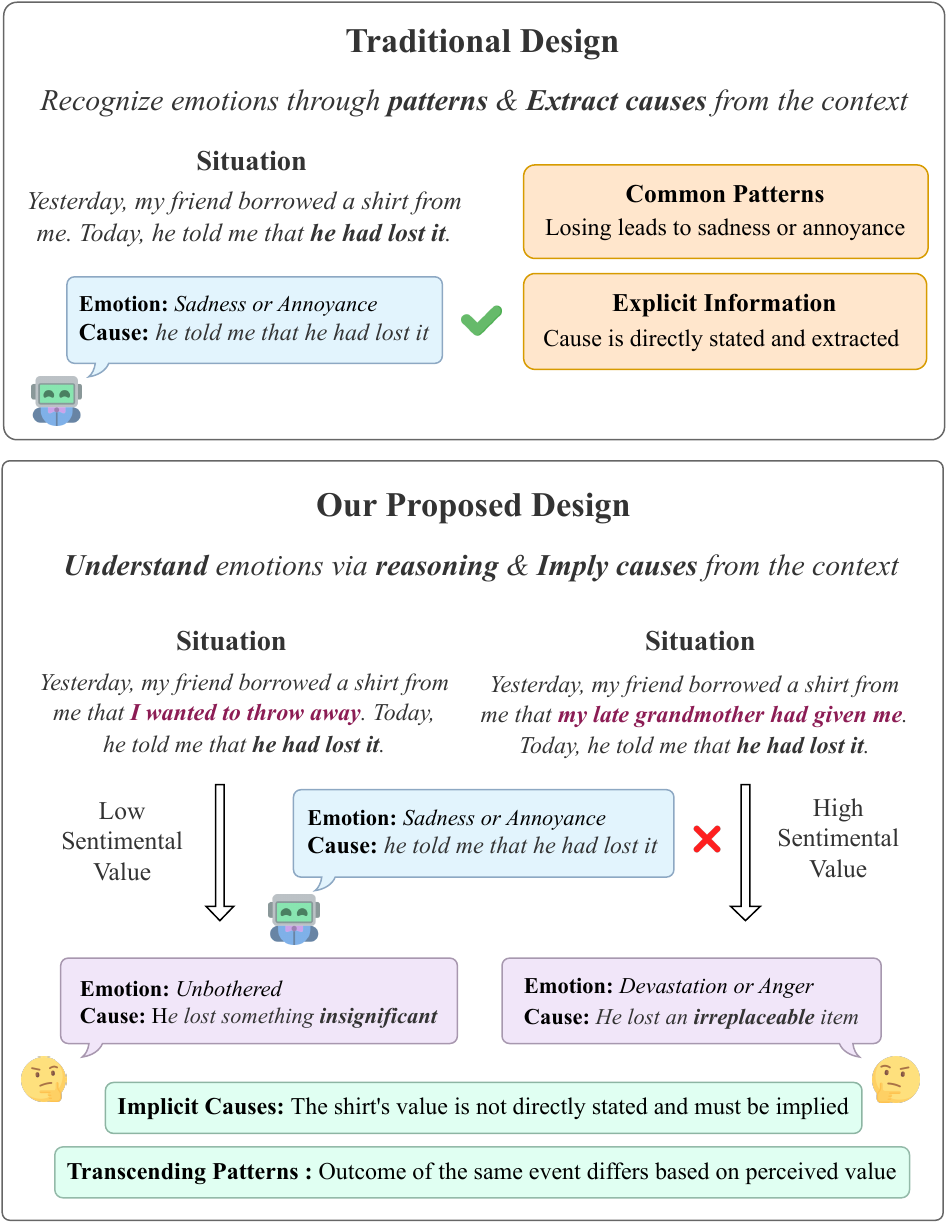}
    \caption{An example of the shortcomings in previous approaches for emotion label and cause recognition and our proposed solution. 
    In this scenario, the perceived value of an object is directly correlated with the person's emotion and its intensity. 
    Rather than extracting part of the context, this perceived value, which serves as the cause for emotions, should be \textit{implied} from the context, increasing the difficulty and practicality of the dataset.
    }
    \label{fig:main_pic}
\end{figure}

Emotional intelligence (EI) enables us to recognize, understand, and manage the thoughts and feelings of ourselves and others \cite{salovey1990emotional}.
It plays a pivotal role in shaping our interpersonal relationships, improving our decision-making, and impacting our overall well-being \cite{schutte2001emotional, schutte2002characteristic, lopes2004emotional}.
Notably, emotionally intelligent systems share similar benefits \cite{reeves1996media}, as they are perceived as more understanding, trustworthy, and engaging \cite{fan2017we, sidner2016engagement}.
These traits are crucial in many areas with widespread applications such as education, customer service, and emotional and mental health support \cite{ivanovic2014emotional, del2021emotional, liu2021towards}.

Recent large language models (LLMs) \cite{bai2023qwen, yang2023baichuan, touvron2023llama, openai2023gpt4} have pushed the boundaries of our expectations regarding their potential capabilities.
However, despite their apparent proficiency in a variety of downstream tasks, such as question answering, and summarization \cite{zhou2023solving, zhong2023agieval}, research on evaluating EI capabilities for LLMs has been limited.
The majority of current benchmarks \cite{huang2023chatgpt, yang2023evaluations, amin2023wide} assess EI through existing datasets for traditional tasks, mainly \textit{Emotion Label and Cause Recognition}.
Yet, these datasets were mainly designed as pattern recognition problems \cite{picard2008toward}, encouraging models to rely on frequent patterns and explicit information \cite{xu2023efficient} rather than implications and reasoning \cite{ghosal-etal-2022-cicero}.
Moreover, EI is not only limited to recognizing emotions and their causes, but also includes the ability to understand emotions and leverage this understanding for thought facilitation and emotion management \cite{maccann2008new}.
We believe the advancing capabilities of LLMs require the development of more comprehensive and challenging benchmarks for EI.
These benchmarks should go beyond conventional tasks to fully evaluate LLMs' understanding, reasoning, and ability to navigate individuals' mental states, encompassing all of the core EI capabilities.
 
An example highlighting these issues is provided in Figure \ref{fig:main_pic}.
Traditional datasets typically contain samples that adhere to common patterns, such as associating 'losing' with 'sadness', and include explicit information guiding the model to extract the cause directly from the context.
However, by simply adding an object's perceived value, the model would need to deduce the individual's mental state in the provided scenario to identify the corresponding emotion and infer its corresponding cause.
 
Towards this end, we propose \textsc{EmoBench}, a theory-based comprehensive EI benchmark for LLM evaluation, consisting of a set of 400 hand-crafted questions, available in English and Chinese.
Our framework draws upon several established psychological theories for EI  \cite{salovey1990emotional, goleman1996emotional, schuller2018age, o2019measurement, rivers2020emotional} and presents an extensive definition for machine EI, covering its essential capabilities: Emotional Understanding (EU) and Emotional Application (EA).
We design emotionally sophisticated scenarios involving multiple individuals and multi-label annotations, encompassing diverse social situations, relationships, and emotional problems. 
In our evaluation, we assess an LLM's ability to accurately \textit{understand} the emotions of the individuals in the scenario and their causes (EU).
We also evaluate whether they can appropriately \textit{apply this understanding}  (EA) to facilitate their thoughts and emotion management and identify the most effective solution within an emotional dilemma (e.g., a family member asking for money when you are facing financial problems yourself).
Our experimental results highlight a considerable gap between the EI capabilities of existing LLMs and humans, with the best-performing LLM (GPT-4) falling short of the average human's performance.

To the best of our knowledge, \textsc{EmoBench} is the first benchmark to propose a comprehensive framework for EI, including assessments of emotional understanding and application.
In line with our work, \citet{Wang2023emotional} and \citet{paech2023eq} also curated similar assessments for EI.
However, their evaluation is limited to Emotional Understanding and is also comparatively limited in scale.
We will publicly release our code and data to facilitate future research on this topic.

\begin{figure*}[t]
    \centering
    \includegraphics[width=\linewidth]{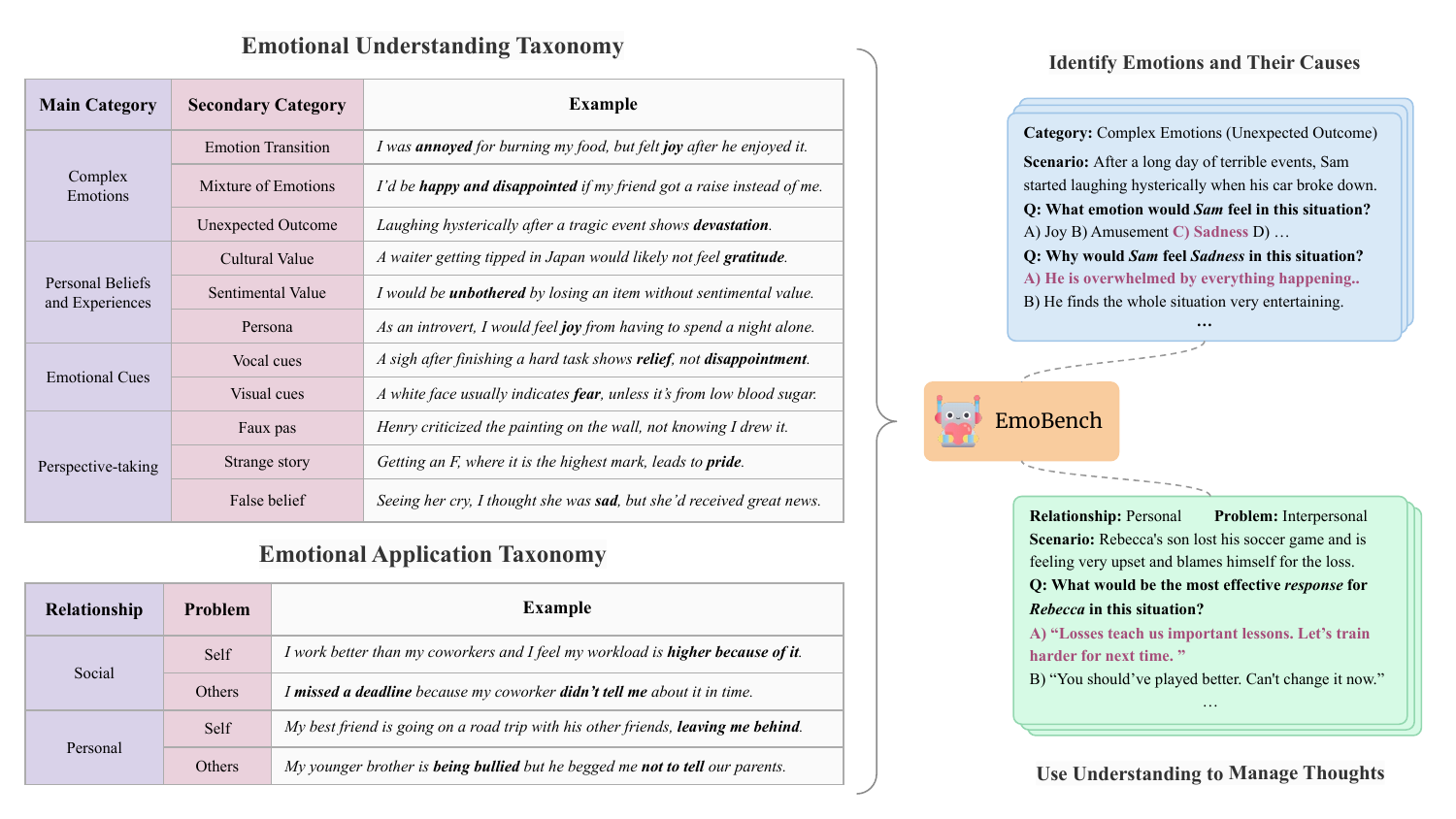}
    \caption{Overview of Our Benchmark (\textsc{EmoBench}).}
    \label{fig:framework}
\end{figure*}

\section{Preliminaries}
\subsection{Definition of Emotional Intelligence}\label{sec:definition}
The term \textit{Emotional Intelligence} was coined and popularized by \citet{salovey1990emotional} as the ability to monitor feelings of our own and understand feelings of others, differentiate between them, and leverage this information to guide our thoughts and actions.
Since then, the rapid progress in psychology research has expanded our understanding of EI, facilitating the rise of new perspectives on EI \cite{bar1997baron, goleman1996emotional,schuller2018age} and improvements upon existing definitions \cite{salovey1990emotional, mayer1999emotional, rivers2020emotional}.
The differences in perspectives and definitions of EI make its assessment a non-trivial task \cite{waterhouse2006multiple}, as the experimental interpretations rely heavily on the adopted definitions and criteria. 
Hence, we must first identify commonalities of existing work and establish a comprehensive definition of machine EI. 

At its core, EI is a unique set of abilities.
Among the most notable definitions, \citet{mayer1999emotional} suggested EI is the ability to perceive, understand, regulate, and express emotions. 
\citet{goleman1996emotional} and \cite{bar1997baron} believed competence in five aspects is indicative of high EI: knowing, recognizing, and managing emotions in self and others, motivating oneself, and building relationships.
In addition, \citet{schuller2018age}'s interpretation of EI involved emotion recognition, adapting emotions to the situation, and leveraging emotional information to solve problems and accomplish goals.

While there are subtle differences among these interpretations, the recurring theme suggests that a comprehensive view of EI revolves around the ability to accurately \textit{understand emotions}, which includes perceiving, identifying, and monitoring emotions, and appropriately \textit{applying this understanding} to accomplish a task (e.g., managing emotions and facilitating our thoughts and decisions).
Hence, we designed our evaluation framework to encompass these two salient dimensions: Emotional Understanding (EU) and Emotional Application (EA).

\subsection{Measures of Emotional Intelligence}\label{sec:measure}
In psychology, EI evaluation is mainly classified into trait and ability measures \cite{ashkanasy2005rumors}. 
Trait measures are commonly assessed through self-report questionnaires and designed to explore how individuals respond to scenarios that evoke emotions\cite{o2019measurement}.
However, self-report assessments are not suitable for evaluating LLMs.
On the other hand, ability measures target individuals' emotional understanding and performance and provide a more theoretical view of EI, and they are more commonly employed for assessing EI \cite{conte2005review}.
Among them, the Mayer–Salovey–Caruso Emotional Intelligence Test (MSCEIT) \cite{mayer2007mayer} and \citet{maccann2008new}'s situational tests for emotion understanding and management (STEU and STEM), have become the most frequently adopted tools in the literature \cite{o2019measurement}.
These measures include sets of meticulously designed multiple-choice questions, with each set targeting a specific EI ability.

\section{\textsc{EmoBench}}  \label{sec:methodology}
We believe EI benchmarks should be comprehensive and transcend general patterns while necessitating deep reasoning and understanding.
Therefore, based on our established definition for machine EI (\S\ref{sec:definition}) and existing tools for EI assessment in psychology (\S\ref{sec:measure}), our framework includes a multi-faceted evaluation of LLMs' emotional \textit{understanding}, while also exploring LLMs' emotional awareness and mentalizing capabilities by analyzing their response to emotional dilemmas and their \textit{application} of emotional understanding.

Figure \ref{fig:framework} presents an overview of \textsc{EmoBench}.
First, through synthesizing several established psychological theories for EI  \cite{salovey1990emotional, goleman1996emotional, rivers2020emotional}, we identified and taxonomized essential capabilities for the established dimensions: Emotional Understanding (EU) and Emotional Application (EA).
Accordingly, based on these taxonomies, we crafted a series of emotionally sophisticated situations involving one to three individuals.

Creating challenging scenarios that involve implications and do not rely on common patterns requires substantial creativity and diversity, which makes manual data collection a non-trivial task.
Therefore, using the designed category descriptions, we initially prompted GPT-4 \cite{openai2023gpt4} to generate example scenarios.
However, while GPT-4 produced the best results in our preliminary experiments among the adopted LLMs, the generated scenarios included explicit mentions of emotion labels and their causes and required minimum reasoning and understanding to reach the correct answer, lacking emotional depth and coverage.
Therefore, we used the generated examples as inspiration to increase our topic diversity and manually crafted the scenarios in our dataset.
Lastly, we annotated each scenario based on each dimension's design and requirements, which we will discuss in the following sections.
For the remainder of this section, the authors who collected and annotated the data will be referred to as workers.

\subsection{Emotional Understanding} 
Emotion Recognition has become a popular research direction in NLP over the past two decades as it is an essential skill for emotionally intelligent machines \cite{picard2001toward}.
There exist several datasets that are commonly used for this task, such as MELD \cite{poria-etal-2019-meld}, DailyDialog \cite{li2017dailydialog}, and GoEmotions \cite{demszky2020goemotions}. 
These datasets mainly provide an emotion-stimulating scenario and a corresponding emotion label for the person involved in the situation (e.g.,  \textit{I broke up with my girlfriend} $\rightarrow$ \textit{Sad}).
Following this trend, an auxiliary task, namely Emotion Cause Recognition \cite{poria2021recognizing}, was proposed to assess whether language models can learn to identify the causes of emotions in addition to their labels in given scenarios (e.g., \textit{I'm getting married soon} $\rightarrow$ \textit{getting married} $\rightarrow$ \textit{Excited}).

There are two fundamental problems with the design of these traditional datasets.
First, previous work considers emotion recognition as a pattern recognition problem \cite{picard2008toward, schuller2018age}, in which models predict the most likely emotion label for the situation based on the observed patterns in the training set.
With this approach, no reasoning or understanding is involved or required to reach the desired output, a trait we believe is necessary for evaluating modern LLMs due to their emerging capabilities. 
Moreover, current datasets for cause recognition are designed as span extraction problems, requiring the cause to be explicitly stated and removing the need for understanding the individual's mental state and reasoning about implications.

However, we believe combining these two tasks lays a solid foundation for assessing emotional understanding.
Hence, while keeping the same format, we create more challenging scenarios in which merely relying on common patterns would not lead to the correct response, and understanding emotional implications and thorough reasoning is necessitated.
Moreover, as many of our designed scenarios involve multiple individuals, our assessment targets understanding the various perspectives of the same situation, which leads to differences in the experienced emotions.

\paragraph{Data Collection and Annotation}\label{sec:EU_method}
Our designed taxonomy for this dimension predominately assesses LLMs' comprehension of four essential categories that are indicative of emotional understanding: \textbf{\textit{complex emotions}}, \textbf{\textit{emotional cues}}, \textbf{\textit{personal beliefs and experiences}}, and individual perspectives (\textbf{\textit{perspective-taking}}).
Each category consists of several sub-categories, targeting its various aspects. More descriptions and examples are provided in Appendix \ref{sec:EU_appendix}.

\begin{figure}[!ht]
    \centering
    \includegraphics[width=\linewidth]{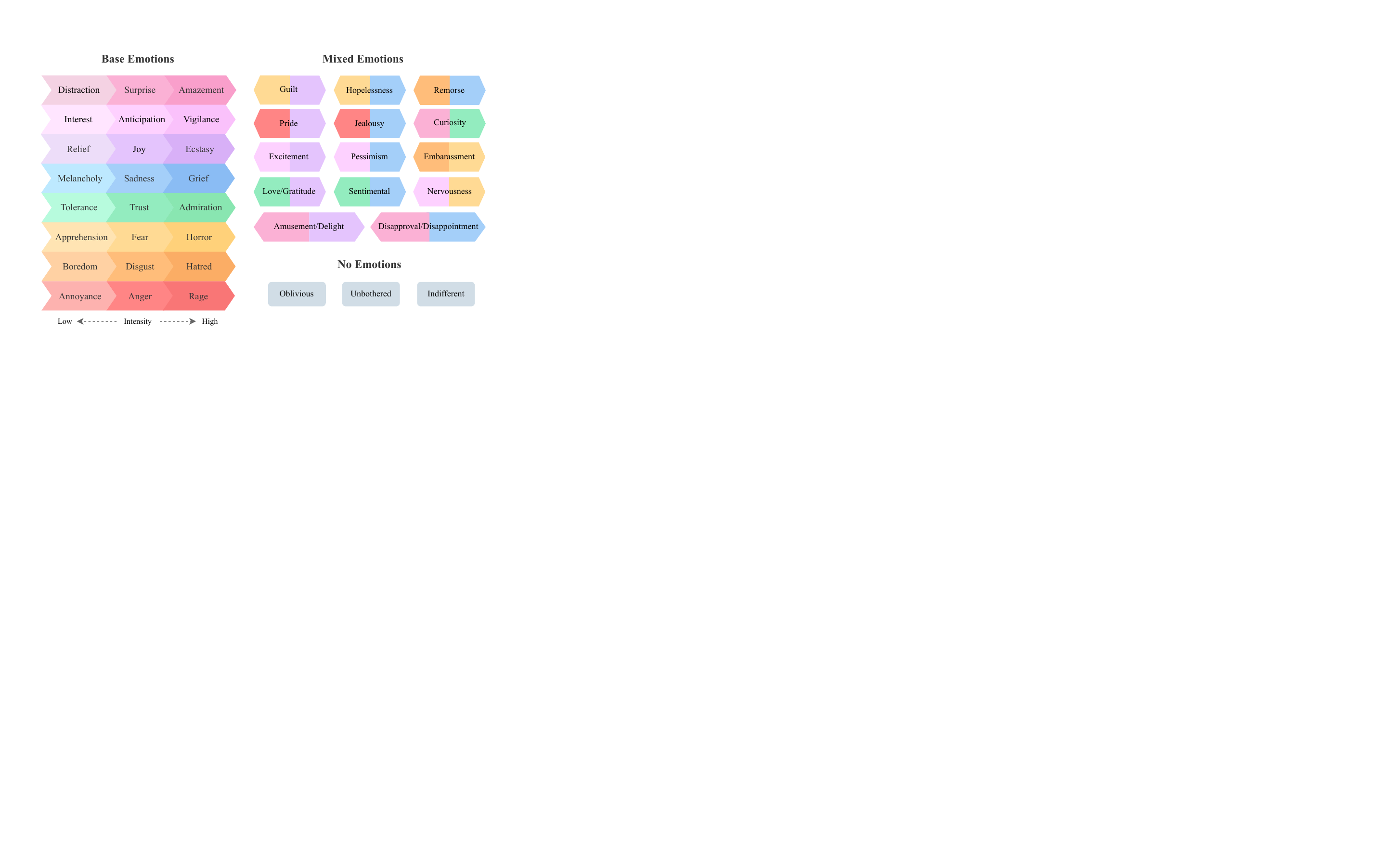}
    \caption{Emotion Taxonomy in \textsc{EmoBench}.}
    \label{fig:emotion_pic}
\end{figure}

Subsequently, in our framework, we need to annotate the labels and causes for the emotions of the people involved in the scenario.
Due to its comprehensive and scalable design, we adopt Plutchik's wheel of emotions \cite{plutchik1982psychoevolutionary} as the foundation of our emotion taxonomy. 
At its core, Plutchik's design involves eight basic emotions with varying intensities, and other emotions are created and labeled as a mixture of these basic emotions.
For instance, the basic emotion \textit{Disgust} could turn into \textit{Boredom} or \textit{Loathing} with low and high intensities, respectively.
It could also mix with \textit{Sadness} to create the feeling of \textit{Remorse}.
This design facilitates the addition of new labels by mixing different emotions and seamless scaling of our taxonomy.
In addition, we aggregate the emotion labels from previous work \cite{ekman1984expression, li2017dailydialog, rashkin2018towards, demszky2020goemotions} to augment our emotion categories, creating a unified and scalable taxonomy (Figure \ref{fig:emotion_pic}).

\begin{figure}[ht]
    \centering
    \includegraphics[width=\linewidth]{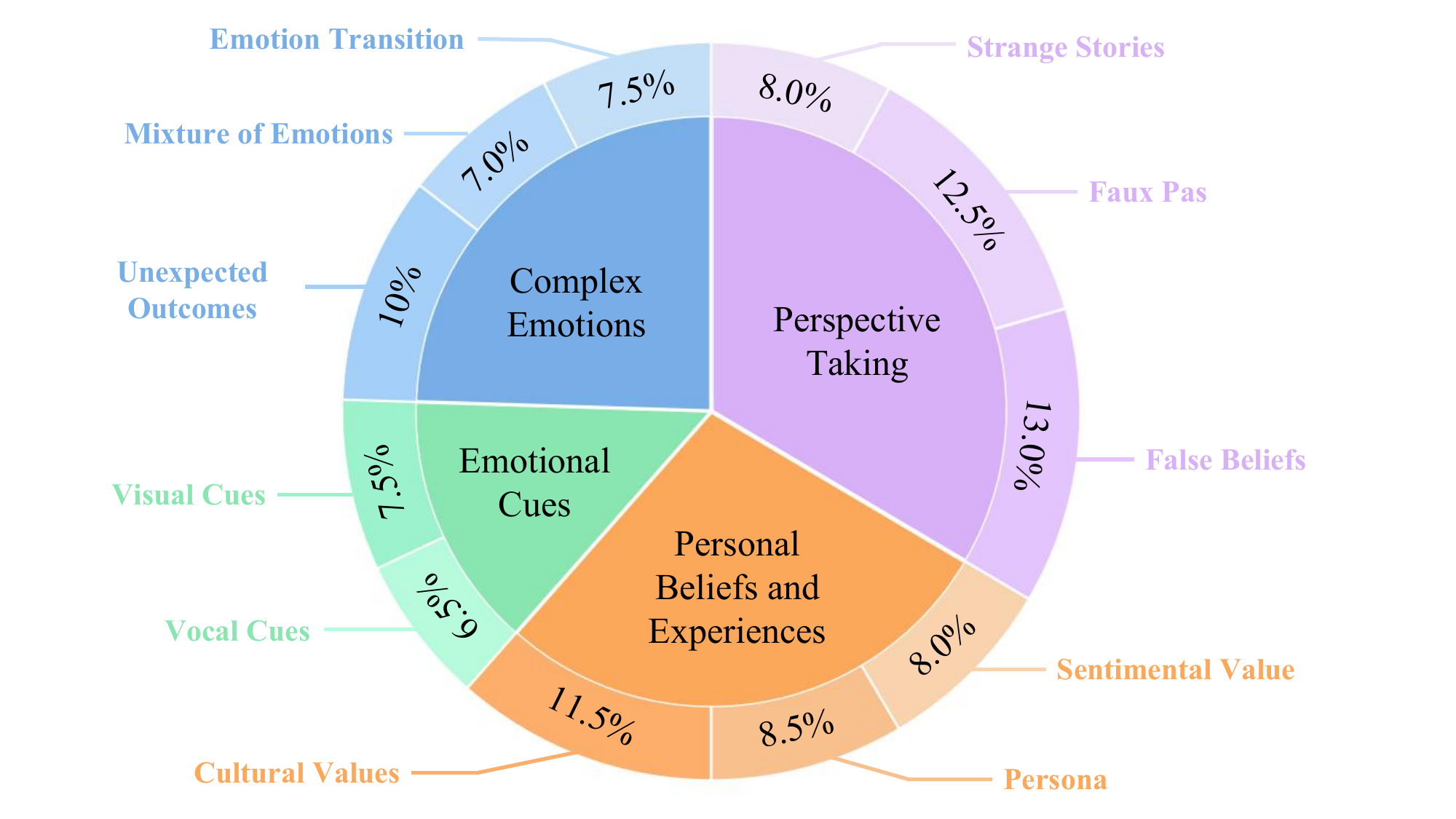}
    \caption{Category Distribution in \textsc{EmoBench}. The main categories are depicted within the chart, and the secondary categories are annotated outside the chart.}
    \label{fig:topic_pic}
\end{figure}

Following the design of our taxonomies, each worker manually created emotionally challenging scenarios and annotated the emotion label and cause for the people involved.
They also created additional labels for emotions and causes to form multiple-choice questions (MCQs).
In our framework, a scenario involving three people would result in three separate MCQs for each individual's emotions and their causes, respectively.
Subsequently, one worker was assigned to translate the MCQ (into English if the original was written in Chinese, and vice versa, based on the worker's language fluency) and, with the addition of two other workers, meticulously review its content to ensure data quality and overall agreement.
In total, we created 121 scenarios involving 1-3 individuals, leading to 200 challenging MCQs.
Figure \ref{fig:topic_pic} shows the corresponding category distributions (emotion distributions are provided in Appendix \ref{sec:emo_dist}).

\subsection{Emotional Application}
Despite emotional understanding being a critical part of EI, it is also essential to analyze how LLMs use this knowledge to facilitate thoughts and manage emotions when faced with emotionally sophisticated problems \cite{goleman1996emotional}. 
Inspired by \citet{maccann2008new}, we propose a novel task for assessing LLM's EI: Emotional Application.
In this task, we aim to evaluate LLMs' proficiency in leveraging their emotional understanding of the individuals' mental states in a given scenario and identifying the most effective course of action or response within an emotional dilemma.

We create our scenario based on different \textbf{\textit{Relationships}} and \textbf{\textit{Problems}}. 
Similar to \citet{zhou2023sotopia}, we only consider two types of relationships in this work: \textit{personal} (e.g., friends, family, romantic partners) and \textit{social} (e.g., boss, teacher, coworkers), and leave more detailed categorizations to future work.
Accordingly, a situation involving these relationships could contain problems that we (\textit{self}) or \textit{others} are facing.
Issues arising from interpersonal conflicts or arguments are also considered problems with \textit{others}.
Lastly, we would prompt the LLM to find the most effective solution to the presented dilemma, which is either an \textit{action} (i.e., what to do?) or a \textit{response} (i.e., what to say?).

\paragraph{Data Collection and Annotation}

Similar to Section \ref{sec:EU_method}, each worker was tasked with designing scenarios based on the generated examples and the assigned categories, and creating multiple plausible solutions to the presented dilemma.
Workers were encouraged to make the MCQ more difficult by introducing implications in the scenario and making all of the choices plausible.
Subsequently, a second worker revised and translated the scenario and choices (English $\rightarrow$ Chinese, and vice versa).  

Given that this could be seen as a subjective task, we assigned the original two workers alongside two new workers to annotate each MCQ and determine its label. 
Inspired by \citet{maccann2008new}, workers were asked to distribute four units of $0.25$ based on their preference as scores for the available choices ($\sum \text{Scores} = 1$). 
For instance, for choices $\{a,b,c,d\}$, if a worker believes choices $a$ and $b$ are both plausible but prefers $a$ over $b$, the annotated score would be $\{0.75,0.25,0,0\}$.
Then, we averaged the scores from all annotators to define the most effective answer for each dilemma.
The inter-annotator agreement using Fleiss' Kappa \cite{fleiss1973equivalence} was $\kappa = 0.852$), indicating excellent agreement and an objectively correct answer for the majority of the collected questions.
Overall, we curated a set of 200 MCQs, with each \textit{relationship-problem-solution} triplet (e.g., social-self-action) containing 25 items.

\section{Experiments}  \label{sec:experiments}
\subsection{Task Formulation}
Our tasks take the form of multiple-choice questions (MCQ). 
For each MCQ in the Emotional Understanding task, we first ask the LLM to identify the individual's emotion and, subsequently, choose the corresponding cause.
In the Emotional Application task, we simply ask the LLM to choose the most effective response/action in the given scenario.
We evaluate LLMs in two settings: zero-shot prompting with task instruction (\textbf{Base}) and with chain-of-thought reasoning (\textbf{CoT}).
Our designed prompts are provided in Appendix \ref{sec:prompts_appendix}.

For our evaluation, we prompt each LLM five times (5-shot) for each MCQ and use majority voting (i.e., the most frequent choice) to determine the LLM's answer. 
Then, we leverage a series of heuristic rules to parse the generated outputs.
Since LLMs have shown to have a bias towards choice ordering \cite{zheng2023large}, we randomly modify the choice ordering three times (4 permutations) and repeat the above process for each new permutation.
Lastly, we calculate and report the average accuracy of the four runs.

\definecolor{lightGreen}{HTML}{d6fae5}
\definecolor{lightBlue}{HTML}{D9E9F7}

\newcommand{\wineu}[1]{{\colorbox{lightGreen}{\color{black}{\textbf{#1}}}}}
\newcommand{\winea}[1]{{\colorbox{lightBlue}{\color{black}{\textbf{#1}}}}}
\begin{table*}[!ht]
  \centering
  \resizebox{2.08\columnwidth}{!}{
    \begin{tabular}{l | c c c c c c c c | c c}
        \toprule
         \textbf{Emotional Understanding Ability} & \multicolumn{2}{c}{\textbf{CE}} & \multicolumn{2}{c}{\textbf{PBE}} & \multicolumn{2}{c}{\textbf{PT}} & \multicolumn{2}{c}{\textbf{EC}}&\multicolumn{2}{|c}{\textbf{Overall}}\\
        \midrule
        \textbf{LLM} & \textbf{EN} & \textbf{ZH} & \textbf{EN} & \textbf{ZH} & \textbf{EN} & \textbf{ZH} & \textbf{EN} & \textbf{ZH}& \textbf{EN} & \textbf{ZH}\\
        \midrule
        \href{https://huggingface.co/01-ai/Yi-6B}{Yi-Chat-6B (Base) }&  16.33 & 20.41 & 12.95 & \wineu{20.54} & 7.84 & 13.43 & 17.86 & 24.11& 12.75 & 18.62 \\
        \href{https://huggingface.co/01-ai/Yi-6B}{Yi-Chat-6B (CoT) }&  12.76 & 17.35 & 10.27 & 12.05 & 8.21 & 11.19 & 20.54 & 16.96& 11.62 & 13.75\\
        \href{https://huggingface.co/THUDM/chatglm3-6b}{ChatGLM3-6B (Base)}& 24.49 & 30.61 & 19.64 & 14.73 & 13.43 & 11.19 & \wineu{30.36} & 37.50 & 20.25 & 20.62 \\ 
        \href{https://huggingface.co/THUDM/chatglm3-6b}{ChatGLM3-6B (CoT)}& 22.96 & 26.53 & \wineu{21.88} & 17.41 & 14.55 & 13.06 & 26.79 & \wineu{38.39}& 20.38 & \wineu{21.12} \\ 
        \href{https://huggingface.co/meta-llama/Llama-2-7b-chat-hf}{Llama2-Chat-7B (Base)} & 13.27 & 13.27 & 9.37 & 9.37 & 13.06 & 4.85 & 10.71 & 5.36 & 11.75 & 8.25  \\ 
        \href{https://huggingface.co/meta-llama/Llama-2-7b-chat-hf}{Llama2-Chat-7B (CoT)} & 8.67 & 7.65 & 5.80 & 4.02 & 6.72 & 10.07 & 2.68 & 0.89 & 6.38 & 6.50\\ 
        \href{https://huggingface.co/baichuan-inc/Baichuan2-7B-Chat}{Baichuan2-Chat-7B (Base)} & \wineu{30.10} & 25.00 & 20.98 & 12.50 & 16.04 & 13.06 & 26.79 & 36.61& 22.38 & 19.12 \\
        \href{https://huggingface.co/baichuan-inc/Baichuan2-7B-Chat}{Baichuan2-Chat-7B (CoT)} & 26.53 & 20.92 & 14.73 & 10.71 & 15.30 & \wineu{17.91} & 22.32 & 22.32  & 18.88 & 17.25\\
        \href{https://huggingface.co/Qwen/Qwen-7B-Chat}{Qwen-Chat-7B (Base)}& 28.06 & \wineu{26.02} & \wineu{21.88} & 16.96 & \wineu{16.42 }& 15.30 & 28.57 & 31.25& \wineu{22.50} & 20.62  \\
        \href{https://huggingface.co/Qwen/Qwen-7B-Chat}{Qwen-Chat-7B (CoT)}& 25.51 & 16.33 & \wineu{21.88} & 15.62 & 15.67 & 13.06 & 26.79 & 25.00& 21.38 & 16.25  \\
        \midrule
        \href{https://huggingface.co/meta-llama/Llama-2-13b-chat-hf}{Llama2-Chat-13B (Base)}&  24.49 & 15.82 & 13.84 & 10.27 & 15.30 & 13.06 & 22.32 & 14.29 &  18.12 & 13.12 \\ 
        \href{https://huggingface.co/meta-llama/Llama-2-13b-chat-hf}{Llama2-Chat-13B (CoT)}&  14.29 & 11.22 & 11.16 & 7.59 & 11.19 & 12.69 & 16.07 & 5.36& 12.62 & 9.88 \\ 
        \href{https://huggingface.co/baichuan-inc/Baichuan2-13B-Chat}{Baichuan2-Chat-13B (Base)}& 34.69 & 37.24 & 24.55 & 19.64 & 18.66 & 20.15 & 33.04 & 37.50 & 26.25 & 26.62  \\
        \href{https://huggingface.co/baichuan-inc/Baichuan2-13B-Chat}{Baichuan2-Chat-13B (CoT)}& 27.55 & 29.08 & 16.07 & 16.07 & 13.81 & 16.79 & 25.00 & 33.93 & 19.38 & 22.00\\
        \href{https://huggingface.co/Qwen/Qwen-14B-Chat}{Qwen-Chat-14B (Base)} & \wineu{46.94} & \wineu{43.37} & \wineu{35.27} & \wineu{30.36} & \wineu{26.12} & 19.40 & \wineu{38.39} & \wineu{41.96} & \wineu{35.50} & \wineu{31.50} \\
        \href{https://huggingface.co/Qwen/Qwen-14B-Chat}{Qwen-Chat-14B (CoT)} & 43.37 & 41.84 & 25.45 & 25.00 & 22.76 & \wineu{21.27} & 33.93 & \wineu{41.96} & 30.12 & 30.25\\
        \midrule
        \href{https://huggingface.co/baichuan-inc/Baichuan2-13B-Chat}{Baichuan2-Chat-53B (Base)}&  43.88 & 46.43 & 31.25 & 25.00 & 25.37 & 25.37 & 49.11 & 50.89&  34.88 & 34.00 \\
        \href{https://huggingface.co/baichuan-inc/Baichuan2-13B-Chat}{Baichuan2-Chat-53B (CoT)}&  41.33 & 57.14 & 28.57 & 26.79 & 25.37 & 11.94 & 45.54 & 53.57 & 33.00 & 33.00\\
        \href{https://huggingface.co/THUDM/chatglm3-6b}{ChatGLM3-66B (Base)}& 47.45 & 42.86 & 30.36 & 25.89 & 26.49 & 29.85 & 50.89 & 54.46& 36.12 & 35.38 \\
        \href{https://huggingface.co/THUDM/chatglm3-6b}{ChatGLM3-66B (CoT)}& 42.35 & 36.73 & 30.80 & 21.43 & 25.00 & 25.37 & 45.54 & 42.86& 33.75 & 29.50\\
        \href{https://openai.com/blog/chatgpt}{GPT 3.5 (Base)}& 41.84 & 30.61 & 33.48 & 18.30 & 21.64 & 22.01 & 44.64 & 45.54& 33.12 & 26.38  \\
        \href{https://openai.com/blog/chatgpt}{GPT 3.5 (CoT)}& 43.88 & 34.69 & 29.46 & 16.96 & 26.49 & 20.52 & 42.86 & 46.43 & 33.88 & 26.62\\
        \href{https://openai.com/gpt-4}{GPT 4 (Base)} & \wineu{72.45}$^\dag$ & 66.84 & \wineu{54.46}$^\dag$ & \wineu{45.09}$^\dag$ & \wineu{50.37}$^\dag$ & \wineu{43.28}$^\dag$ & 70.54 & \wineu{75.89}$^\dag$ &\wineu{59.75}$^\dag$  & \wineu{54.12}$^\dag$ \\
        \href{https://openai.com/gpt-4}{GPT 4 (CoT)} & 68.88 & \wineu{68.37}$^\dag$ & 53.13 & 43.30 & 49.25 & 41.79 & \wineu{71.43}$^\dag$ & 63.39  & 58.25 & 51.75\\
        \midrule
        \textbf{Random} & \multicolumn{2}{c}{2.04} & \multicolumn{2}{c}{3.12} & \multicolumn{2}{c}{3.36} & \multicolumn{2}{c}{1.79}& \multicolumn{2}{|c}{2.62}\\
        \textbf{Majority} & \multicolumn{2}{c}{16.33} & \multicolumn{2}{c}{8.93} & \multicolumn{2}{c}{14.29} & \multicolumn{2}{c}{13.43}& \multicolumn{2}{|c}{11.5}\\
        \bottomrule
    \end{tabular}
    }
  \caption{Evaluation Results for \textsc{EmoBench}'s \textit{Emotional Understanding} (accuracy \%). 
  The best results for LLMs with similar sizes are highlighted in \wineu{Bold}, with the best overall results marked by $^\dag$. \textbf{CE}, \textbf{PBE}, \textbf{PT}, \textbf{EC} indicate Complex Emotions, Personal Beliefs and Experience, Perspective Taking, and Emotional Cues, respectively.
  }
  \label{table:eu_results}

\end{table*}

\begin{table*}[!ht]
  \centering
  \resizebox{2.08\columnwidth}{!}
  {
    \begin{tabular}{l | c c c c c c c c | c c}
        \toprule
         {\textbf{Relationship-Problem}} & \multicolumn{2}{c}{\textbf{Personal-Self}} & \multicolumn{2}{c}{\textbf{Personal-Others}} & \multicolumn{2}{c}{\textbf{Social-Self}} & \multicolumn{2}{c}{\textbf{Social-Others}}&\multicolumn{2}{|c}{\textbf{Overall}}\\
        \midrule
        {\textbf{LLM}} & \textbf{EN} & \textbf{ZH} & \textbf{EN} & \textbf{ZH} & \textbf{EN} & \textbf{ZH} & \textbf{EN} & \textbf{ZH}& \textbf{EN} & \textbf{ZH}\\
        \midrule
        \href{https://huggingface.co/01-ai/Yi-6B}{Yi-Chat-6B (Base) }&  50.50 & 54.00 & 40.00 & 49.00 & 50.50 & 54.00 & 48.00 & 49.50 & 47.25 & 51.62 \\
        \href{https://huggingface.co/01-ai/Yi-6B}{Yi-Chat-6B (CoT) }&  47.00 & 45.50 & 46.00 & 37.50 & 42.50 & 43.00 & 40.50 & 36.50 & 44.00 & 40.62\\
        \href{https://huggingface.co/THUDM/chatglm3-6b}{ChatGLM3-6B (Base)}& 62.00 & 48.00 & 55.00 & 47.50 & 51.50 & 47.00 & \winea{54.00} & 44.50 & \winea{55.62} & 46.75 \\ 
        \href{https://huggingface.co/THUDM/chatglm3-6b}{ChatGLM3-6B (CoT)}& 61.00 & \winea{54.50} & 52.00 & \winea{56.00} & 52.00 & 52.50 & 46.50 & \winea{52.00} & 52.88 & \winea{53.75} \\ 
        \href{https://huggingface.co/meta-llama/Llama-2-7b-chat-hf}{Llama2-Chat-7B (Base)} & 58.50 & 44.50 & \winea{55.50} & 36.00 & 45.00 & 34.00 & 41.50 & 42.50 & 50.12 & 39.25  \\ 
        \href{https://huggingface.co/meta-llama/Llama-2-7b-chat-hf}{Llama2-Chat-7B (CoT)} & 37.50 & 29.00 & 29.50 & 25.50 & 25.50 & 30.50 & 35.00 & 24.00 & 31.88 & 27.25\\ 
        \href{https://huggingface.co/baichuan-inc/Baichuan2-7B-Chat}{Baichuan2-Chat-7B (Base)} &  59.50 & 48.50 & 52.00 & 38.00 & 48.50 & 47.50 & 50.00 & 44.00 & 52.50 & 44.50 \\
        \href{https://huggingface.co/baichuan-inc/Baichuan2-7B-Chat}{Baichuan2-Chat-7B (CoT)} & 53.50 & 49.00 & 44.00 & 48.00 & 47.50 & 41.00 & 49.50 & 43.00 & 48.62 & 45.25\\
        \href{https://huggingface.co/Qwen/Qwen-7B-Chat}{Qwen-Chat-7B (Base)}& \winea{62.50} & 44.00 & 50.50 & 49.00 & \winea{55.50} & 51.50 & 50.00 & 42.00 & 54.62 & 46.62  \\
        \href{https://huggingface.co/Qwen/Qwen-7B-Chat}{Qwen-Chat-7B (CoT)}& 49.00 & 53.50 & 40.50 & 53.50 & 50.50 & \winea{55.00} & 36.50 & 48.00 & 44.12 & 52.50  \\
        \midrule
        \href{https://huggingface.co/meta-llama/Llama-2-13b-chat-hf}{Llama2-Chat-13B (Base)}&  68.00 & 55.00 & 53.50 & 45.50 & 53.50 & 55.50 & 48.50 & 46.50 & 55.88 & 50.62 \\ 
        \href{https://huggingface.co/meta-llama/Llama-2-13b-chat-hf}{Llama2-Chat-13B (CoT)}&  48.00 & 40.00 & 34.00 & 32.50 & 35.00 & 33.00 & 34.00 & 29.00 & 37.75 & 33.62 \\ 
        \href{https://huggingface.co/baichuan-inc/Baichuan2-13B-Chat}{Baichuan2-Chat-13B (Base)}& 52.00 & 51.50 & 52.00 & 51.50 & 52.00 & \winea{58.00} & \winea{58.50} & \winea{58.00} & 53.62 & 54.75  \\
        \href{https://huggingface.co/baichuan-inc/Baichuan2-13B-Chat}{Baichuan2-Chat-13B (CoT)}& 52.50 & 46.50 & 51.50 & 43.50 & 47.50 & 48.50 & 52.50 & 42.00 & 51.00 & 45.12\\
        \href{https://huggingface.co/Qwen/Qwen-14B-Chat}{Qwen-Chat-14B (Base)} & \winea{74.00 }&\winea{ 69.00} & \winea{54.00} & 56.50 & \winea{60.50} & 56.50 & 53.50& 50.50 & \winea{60.50 }& \winea{58.12} \\
        \href{https://huggingface.co/Qwen/Qwen-14B-Chat}{Qwen-Chat-14B (CoT)} & 45.50 & 62.50 & 42.00 & \winea{58.00} & 47.50 & 56.50 & 38.00 & 55.00 & 43.25 & 58.00\\
        \midrule
        \href{https://huggingface.co/baichuan-inc/Baichuan2-13B-Chat}{Baichuan2-Chat-53B (Base)}&  43.88 & 46.43 & 31.25 & 25.00 & 25.37 & 25.37 & 49.11 & 50.89&  34.88 & 34.00 \\
        \href{https://huggingface.co/baichuan-inc/Baichuan2-13B-Chat}{Baichuan2-Chat-53B (CoT)}&  41.33 & 57.14 & 28.57 & 26.79 & 25.37 & 11.94 & 45.54 & 53.57 & 33.00 & 33.00\\
        \href{https://huggingface.co/THUDM/chatglm3-6b}{ChatGLM3-66B (Base)}& 71.00 & 65.00 & 59.50 & 53.50 & 65.50 & 64.00 & 66.00 & 54.00 & 65.50 & 59.12\\
        \href{https://huggingface.co/THUDM/chatglm3-6b}{ChatGLM3-66B (CoT)}&69.00 & 62.50 & 59.00 & 57.00 & 65.00 & 64.00 & 59.50 & 57.00 & 63.12 & 60.12\\
        \href{https://openai.com/blog/chatgpt}{GPT 3.5 (Base)}& 64.50 & 57.00 & 61.00 & 57.00 & 60.50 & 53.00 & 59.50 & 56.00 & 61.38 & 55.75 \\
        \href{https://openai.com/blog/chatgpt}{GPT 3.5 (CoT)}&  67.00 & 62.50 & 61.50 & 61.00 & 62.50 & 53.00 & 58.50 & 53.00 & 62.38 & 57.38\\
        \href{https://openai.com/gpt-4}{GPT 4 (Base)} & \winea{79.50}$^\dag$ & \winea{75.50}$^\dag$ & 78.50 & \winea{82.50}$^\dag$ & 73.50 & \winea{70.50}$^\dag$ & 70.50 & 66.50 & 75.50 &\winea{ 73.75}$^\dag$ \\
        \href{https://openai.com/gpt-4}{GPT 4 (CoT)} & 74.50 & \winea{75.50}$^\dag$ & \winea{80.00}$^\dag$ & 80.50 & \winea{74.00}$^\dag$ & 70.00 & \winea{75.00}$^\dag$ & \winea{68.00}$^\dag$ & \winea{75.88}$^\dag$ & 73.50\\
        \midrule
        \textbf{Random} & \multicolumn{2}{c}{31.00} & \multicolumn{2}{c}{22.5} & \multicolumn{2}{c}{23.5} & \multicolumn{2}{c}{23.5}& \multicolumn{2}{|c}{24.12}\\
        \textbf{Majority} & \multicolumn{2}{c}{32.00} & \multicolumn{2}{c}{36.00} & \multicolumn{2}{c}{36.0} & \multicolumn{2}{c}{44.0}& \multicolumn{2}{|c}{37.0}\\
        \bottomrule
    \end{tabular}
    }
  \caption{Evaluation Results for \textsc{EmoBench}'s \textit{Emotional Application} (accuracy \%). 
  The best results for LLMs with similar sizes are highlighted in \winea{Bold}, with the best overall results marked by $^\dag$. 
  }
  \label{table:ea_results}

\end{table*}

\subsection{Baselines}
In our experiments, we adopt a range of recent widely-used LLMs with promising performance on existing benchmarks \cite{zhang2023safetybench}. 
For close-sourced LLMs (accessible through APIs\footnote{\url{https://api.openai.com/v1/chat/completions}}), we evaluate OpenAI's \textbf{GPT 4} (gpt-4) and \textbf{GPT 3.5} (gpt-3.5-turbo) \cite{openai2023gpt4}, \textbf{ChatGLM 3 (66B)} \cite{du2022glm, zeng2022glm}, and \textbf{Baichuan 2 (53B)} \cite{yang2023baichuan}.
For open-source LLMs, we experimented with \textbf{Llama 2} (7B and 13B; \cite{touvron2023llama}), \textbf{Baichuan 2} (7B and 13B), \textbf{Qwen} (7B and 14B; \cite{bai2023qwen}), \textbf{ChatGLM 3 (6B)}, and \textbf{Yi (6B)}\footnote{\url{https://github.com/01-ai/Yi}}.
Following \citet{ismayilzada-etal-2023-crow}, we also include \textbf{Random} choice and \textbf{Majority} (i.e., choosing the most frequent choice) as baselines.

\subsection{Implementation Details}
For Llama-based LLMs, we used the default generation hyperparameters (top-p sampling with $p=0.9$ and temperature $=0.6$).
For others, we directly employed their pre-defined interfaces, either through their online API or the \textsc{chat} function in the Transformers library\footnote{\url{https://github.com/huggingface/transformers}}.
All of our experiments were run on single A100 80GB GPUs.

\section{Results and Findings}\label{sec:results}
Our obtained results are provided in Tables \ref{table:eu_results} and \ref{table:ea_results}.
Overall, \textbf{GPT-4 significantly outperformed the other LLMs in both tasks}.
In general, all LLMs demonstrated better accuracy than random chance.
However, in the EU task, several of the smaller models had worse performance than simply choosing the most frequent choice.
An interesting finding in our experiments was that \textbf{requiring LLMs to reason step-by-step generally had little to no improvements}, even hindering the performance for smaller models (particularly <14B). We will further investigate this issue in \cref{sec:err_analysis}.

Notably, \textbf{the task's language did not have a significant impact on the performance}, with all LLMs (excluding Yi and ChatGLM-6B) performing slightly better in English, which we believe could be due to data distributions in their training data.
This could also explain why Chinese-based LLMs (e.g., Yi) outperform their English-based counterparts, such as LLama 2 (7B), in the Chinese subset of \textsc{EmoBench} despite having a similar size.
However, as we do not have access to the LLMs' pertaining data, we cannot claim any correlations between their training data and performance on our benchmark.
Moreover, \textbf{the performance consistently improved with increased parameters}, which is consistent with previous findings on LLM scaling \cite{brown2020language}.

\begin{figure*}[!ht]
    \centering
    \includegraphics[width=\linewidth]{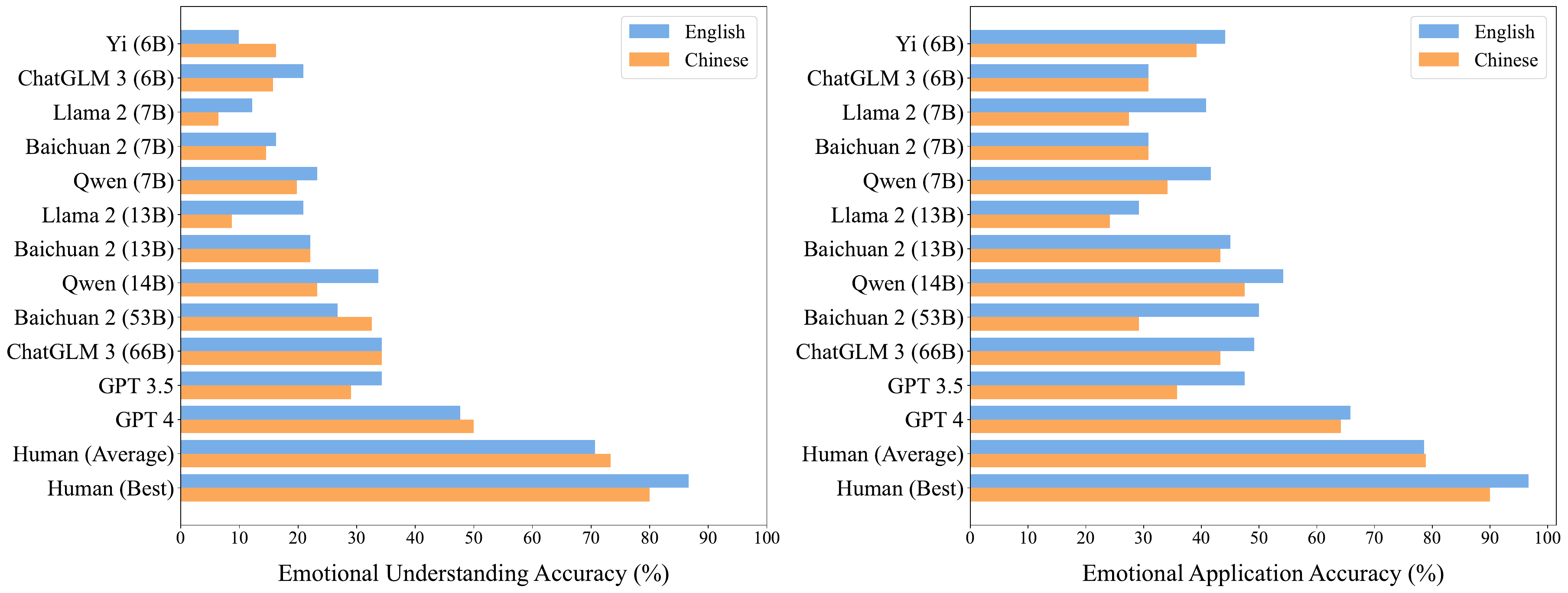}
    \caption{Results on the \textsc{EmoBench} subset used in the human evaluation.}
    \label{fig:human_res}
\end{figure*}

\textbf{All LLMs found emotional understanding considerably more challenging than its application}.
We believe this is due to several reasons.
Contrary to the EA task, the EU samples require LLMs to correctly answer two questions (the emotion and its cause), which itself serves as a bigger challenge.
This is also indicated by the results from the \textbf{Random} and \textbf{Majority} baselines.
Moreover, evidenced by differences in their designs, the EU questions aimed to portray situations that included various implications and outcomes for frequent patterns.
However, our design of EA samples was still prone to including such patterns as with this task, our main goal was to present a novel evaluation of LLMs' awareness and management when faced with emotional dilemmas.
Hence, the difficulty of the EA task would naturally be much lower.

As shown in Table \ref{table:eu_results}, the LLMs found specific categories with the EU task more challenging than the others.
Mainly, all LLMs struggled with MCQs regarding \textbf{Perspective Taking (PT)}, which has also been shown in relevant tasks (e.g., ToM \cite{ullman2023large}) that require this mentalizing ability.
Similarly, LLMs found it difficult to understand the nuances regarding personal traits, sentimental values, and cultural values.
Within the EA task (Table \ref{table:ea_results}), each LLM had varying performances in different types of relationships and problems.
In general, LLMs perceived solving the self's social problems as more challenging among the studied dimensions.

\definecolor{mygreen}{HTML}{14ad56}
\definecolor{myred}{HTML}{cc2424}
\definecolor{myblue}{HTML}{3079BA}
\begin{table*}[!ht]
  \centering {
    \begin{tabular}{l | l}
        \toprule
        \rowcolor{lightGreen}
         \multicolumn{2}{c}{\textbf{Emotional Understanding (EU)}} \\
        \midrule
        \textbf{Scenario} & Despite countless rejections, Susan kept pursuing her goal of finding a job. Today, she \\
        ~ & \textbf{\color{myblue}finally received an offer} and \textbf{\color{myblue}started crying}. At this moment, her mom walked into the \\ ~ & room and ran to embrace her after seeing her cry. \\ 
         \textbf{Question} & Susan's mom feels ...?\hspace{0.1cm} (a) Relief \hspace{0.05cm}\textbf{\color{mygreen}(b) Nervousness} \hspace{0.05cm} (c) Anger \hspace{0.05cm} (d) Delight\\
         \textbf{LLM} & \textbf{\color{myred}(d) Delight}  $\rightarrow$ Her daughter was finally able to get a job offer.\\
        \midrule
        \textbf{Scenario} & I have \textbf{\color{myblue}acrophobia}. Today, the elevator was out of order, and I had to take the stairs to \\
         ~ & reach my office on the \textbf{\color{myblue}last floor of a tall building}. I considered it a \textbf{\color{myblue}nice little exercise}. \\ 
         \textbf{Question} & I feel ...?\hspace{0.1cm} \textbf{\color{mygreen}(a) Excitement} \hspace{0.05cm} (b) Fear \hspace{0.05cm}(c) Disapproval \hspace{0.05cm} (d) Nervousness\\
         \textbf{LLM} & \textbf{\color{myred}(b) Fear} $\rightarrow$ I was afraid of taking the stairs due to my acrophobia.\\
        \midrule
        \textbf{Scenario} & I have a teacher who gives the F grade as the \textbf{\color{myblue} highest mark} since he believes F stands \\
        ~&  for Full.  I just got the results from yesterday's hard test and saw that \textbf{\color{myblue}he gave me an F}.\\
         \textbf{Question} & In this situation, I feel ...?\hspace{0.1cm} (a) Disappointment \hspace{0.05cm} (b) Nervousness \hspace{0.05cm}\textbf{\color{mygreen}(c) Pride \hspace{0.05cm}} (d) Sadness\\
         \textbf{LLM} & \textbf{\color{myred}(a) Disappointment}  $\rightarrow$  I failed the difficult test\\
        \toprule
        \rowcolor{lightBlue}
        \multicolumn{2}{c}{\textbf{Emotional Application (EA)}} \\
        \midrule
        \textbf{Scenario} &  Peter's \textbf{\color{myblue}best friend jokingly tells him} that he is the reason why their group of friends\\
        & \textbf{\color{myblue}keep losing at video games}. \\
         \textbf{Question} & In this situation, what is the most effective response for Peter?\\
         ~& (a) "I thought it was just a game, not a battle to death" \\
         ~& (b) "Maybe if you would explain the strategies better \\
         ~& \textbf{\color{mygreen} (c) "Well, you're not exactly a pro either"} \\
         ~& (d) "I'm sorry. I'll try to improve my skills so we can win"  \\
         \textbf{LLM} & \textbf{\color{myred} (d)} $\rightarrow$ it shows accountability and a willingness to take action to improve the situation.\\
        \bottomrule
    \end{tabular}}
  \caption{Common observed mistakes in our experiments. 
  LLM represents the studied LLMs' general response.
  {\color{mygreen} Green} and {\color{myred} Red} indicate the correct answer and LLM's answer, respectively.
  The content is summarized for brevity.
  }
  \label{table:case_study}
\end{table*}

\section{Comparison with Human Performance}
To obtain a baseline for human EI, we recruited participants through online surveys to complete our EI test.
More information on our recruitment process, quality control, and participant demographics are provided in Appendix \ref{sec:human_stats}.
In total, we recruited 48 participants and allocated an equal number of participants to each language-task evaluation pair.
Subsequently, for each group, we randomly sampled 30 MCQs from \textsc{EmoBench} that were not included in the initial screening process.

As shown in Figure \ref{fig:human_res}, our human participants outperformed the LLMs on both tasks.
Notably, although GPT-4, the top-performing LLM, came close to the average human performance, particularly in the EA task, it still fell short of surpassing individuals with higher emotional intelligence, highlighting a significant gap in current LLMs.

\section{Error Analysis}\label{sec:err_analysis}
To provide a qualitative view of LLMs' performance on our benchmark, we analyzed LLMs' generated reasoning through CoT. 
We believe appropriate reasoning for our tasks would involve traversing the events within the provided scenario and following the transitions in the individual's emotions, demonstrating an understanding of their mental state and the situation's implications.
However, our analysis showed that LLMs' reasoning mainly involved analyzing the provided choices and evaluating the validity of each choice.
While this could be an effective strategy for filtering out the wrong responses, this form of reasoning may overlook the nuanced emotional awareness and considerations involved in human decision-making, which are pivotal parts of EI.

We observed that LLMs' step-by-step reasoning occasionally led to \textbf{changes in the topic} (e.g., turning to a detailed discussion on the necessity of being empathetic in modern society when faced with a scenario about supporting a loved one within an emotional dilemma) or \textbf{refusal to answer} (stating that none of the options are correct).
Such errors were considerably less common in larger models (>50B), which is indicated by the smaller gaps between their performance with and without CoT (Tables \ref{table:eu_results} and \ref{table:ea_results}).
However, these results are expected as more reliable reasoning capabilities emerge when the parameters are scaled above certain thresholds \cite{wei2022chain}.

Moreover, we present several examples of common mistakes made by LLMs in Table \ref{table:case_study}.
For EU questions, LLMs tend to make mistakes mainly by having \textbf{misassumptions} (e.g., a person walking in the door would not immediately know what is going on), and \textbf{incorrect reasoning} (e.g., having a phobia would not necessarily lead to fear or getting an F is not a failure when its the highest score). 
We believe these errors mainly occur due to LLMs' lack of emotional understanding, such as \textbf{weak perspective taking} (as shown in Table \ref{table:eu_results}) and \textbf{reliance on frequent patterns} for reasoning.
With EA questions, LLMs' answers mainly exhibited a preference for \textbf{more general solutions}, disregarding the relationship between individuals, which is an important factor in determining their emotions and subsequent responses.
For instance, while the best course of action when facing criticism may be taking accountability, gentle humor would be a more suitable response to a friend's simple tease as it shows better emotional regulation and awareness.

\vspace{-0.2cm}
\section{Conclusion and Future Work}
\vspace{-0.1cm}
In this paper, we introduced \textsc{EmoBench}, a theory-based, comprehensive, and challenging set of 400 hand-crafted MCQs, including emotionally sophisticated scenarios, for assessing Emotional Intelligence (EI) in Large Language Models through its two salient dimensions: Emotional Understanding and Emotional Application.
Our results revealed that existing LLMs struggle with emotional intelligence (mainly understanding), and there is still a considerable gap between the best-performing LLM in our study and the average human.

We hope that by facilitating EI evaluation, \textsc{EmoBench} can encourage research on emotionally intelligent LLMs, leading to LLMs that are more capable of understanding emotions and applying this understanding in many promising tasks, such as emotional and mental health support \cite{sabour2022chatbots}. 
In addition, we plan to augment \textsc{EmoBench} with more data, exploring the more fine-grained features.

\vspace{-0.2cm}
\section{Limitations}
\vspace{-0.1cm}
with \textsc{EmoBench}, we aimed to ensure high annotation quality and difficulty with our curated samples, which required intensive labor and manual supervision, and thus, compared to existing benchmarks for other tasks, our dataset is limited in scale.
Given our resources, we were only able to collect data in English and Chinese. 
We believe translating our data to other languages could reveal more insights into their seemingly intelligent behavior.

In addition, our benchmark is limited to a single modality (text) as most of the recent prevalent LLMs are text-based.
However, many psychological tests for emotional intelligence (e.g., MSCEIT; \citet{mayer2007mayer}), include assessments of various modalities, such as the individual's tone and facial features. 
Moreover, while we did not directly include samples from GPT-4, we leveraged its generated examples to inspire our MCQs, which might have introduced a bias in our benchmark.
With future improvements in LLMs, we will continue exploring different dimensions of EI and augment our benchmark accordingly. 

In our evaluation, we acknowledge that the choice of prompts could have significantly influenced the LLMs' performance.
However, despite our emphasis on prompt design, we cannot claim our prompts were optimal, and thus, the experimental results are not indicative of LLMs' peak performance in EI.
Moreover, we only experimented with chain-of-thought reasoning to augment the output, which future work could expand upon and propose new reasoning techniques that better apply to emotional scenarios.

Emotional intelligence is still an abstract concept in psychology and our view on it may change with developing research.
Similarly, emotions are not objective, and individual responses to the same situation could vary significantly. 
We strived to design our scenarios and choices in a manner that would only require a general and commonsensical understanding of emotions.
The trade-off here, particularly for designing scenarios for emotional application, was that we could only include scenarios that all the annotators had experienced to ensure reliable annotation, limiting the scope of the topics and relationships covered.

Furthermore, to address the issues with subjectivity, we designed our MCQs to have only one objectively correct answer.
This is more straightforward for the EU questions, as a golden label can be directly defined for the emotions based on the taxonomy. In addition, four different workers checked and agreed upon these golden labels along with the designed causes, suggesting that all the workers found the labels for emotions and corresponding causes to be objectively the only correct choice among the provided choices.
For EA, to reduce the effect of subjectivity, while we create choices that could all be plausible, we require one choice to be clearly more effective and applicable than others. In addition, in cases where two plausible choices are equally favorable, we modify one of the choices to be a viable action in general circumstances while being impractical in the given situation. As shown by our high human annotator agreement in this task (Fleiss’ Kappa = 0.852), we can assume that the proposed evaluation is substantially objective since multiple annotators were able to agree on one correct answer. 

We did not study the effect of more fine-grained personal traits (e.g., detailed experiences, characteristics, and language expression) on the experienced emotions, as we found it outside of our scope.
For instance, during a conflict or confrontation, a person who deals with issues by making jokes may not experience the same level of anger as a serious individual.
We believe future work could explore augmenting our benchmark with more cases and study the effects of these more fine-grained traits.

\section*{Ethical Considerations}
We emphasize that our evaluation is concerned with the perceived view of emotional intelligence, aiming to explore the limitations of existing LLMs through novel and challenging tasks.
In this work, while our proposed definition includes the ability to understand emotions and apply this understanding to manage emotions, we do not claim nor believe that LLMs are capable of possessing or simulating emotions. 
With our experiments, we demonstrated that LLMs still rely on frequent patterns to indicate signs of understanding.
In addition, despite not having emotions, we found that LLMs can capitalize on their seen patterns to show apparent signs of emotional sense and awareness, which is in line with previous research on LLMs' commonsense \cite{sap2019atomic} and morality \cite{jiang2021can}. 

\section*{Acknowledgements}
This work was supported by the National Science Foundation for Distinguished Young Scholars (No. 62125604) and the NSFC Key Project (No.61936010). This work was also supported by Tsinghua Precision Medicine Foundation, Tsinghua University - Beijing Tsingshang Architectural Decoration Engineering Co., Ltd. Joint Institute for Smart Scene Innovation Design.
\bibliography{anthology,custom}
\bibliographystyle{acl_natbib}

\appendix
\pagestyle{fancy}

\section{Scenario Taxonomy}\label{sec:EU_appendix}
\subsection{Complex Emotions} 
Understanding complex emotions is an essential part of emotion understanding \cite{rivers2020emotional}. 
In our framework, we include three categories that cover the essential aspects of complex emotions:
\begin{itemize}
    \item \textbf{Emotion Transition}: In response to different events, our emotions are subject to change. 
    To assess whether LLMs can reason about such transitions in one's emotions, we create scenarios in which the individual's emotion changes based on the turn of events. 
    \begin{quote}
    A mother who is \textit{annoyed} about ruining the food, would be \textit{delighted} when their child enjoys and compliments it.
    \end{quote}
    \item \textbf{Mixture of Emotions}:  while previous work mainly annotates each sample with a single emotion label \cite{li2017dailydialog, rashkin2018towards}, many individuals tend to experience a combination of emotions in various situations.
    Such emotions could be of the same (e.g., happy and excited) or the opposite (e.g., sad yet relieved) polarities.
    Hence, we designed scenarios in which the individual feels a mixture of emotions.
    \begin{quote}
    If two friends, Annie and Mark, participate in the same competition and Annie gets first place, then Mark would be \textit{happy} and \textit{proud} for his friend's accomplishment while being \textit{disappointed} for his loss.
    \end{quote}
    \item \textbf{Unexpected Outcome}: Inspired by \citet{dyck2001autism}, we create scenarios in which the conclusion contradicts explicit commonsense and expected reactions.
    We believe this is crucial in assessing whether LLMs are reliant on patterns to understand emotions, as these scenarios involve reactions that are uncommon for displaying the emotion in the corresponding scenarios.
    \begin{quote}
    If Jamie has had a bad day full of misfortune and bad luck, and finally starts laughing hysterically after dropping his ice cream, his laughter shows \textit{frustration}, not \textit{amusement}.
    \end{quote}
\end{itemize}

\subsection{Personal Beliefs and Experiences} 
To have a deep understanding of one's emotions, we need to recognize how their beliefs and values among past experiences and appraisals could impact the emotions they experience \cite{rivers2020emotional}.
To assess this, we designed three categories that aim to evaluate LLM's comprehension of how individual's \textit{Cultural Values}, \textit{Sentimental Values}, and personal experiences and traits (namely \textit{Persona}) could affect their reaction to certain events.
\begin{itemize}
    \item \textbf{Cultural Values}: In these scenarios, we aim to assess whether LLMs are capable of understanding how an individual's reaction to the same event could vary based on their cultural values and background \cite{rivers2020emotional}.
    Consider the following situation. 
    Anna is brought up in a culture where being late is considered rude. However, Jonah's culture does not put a great emphasis on punctuality.
    \begin{quote}
    If Anna is late to a meeting with Jonah, she would be \textit{embarrassed} and apologetic, while Jonah would be \textit{unbothered}.
    \end{quote}
    \item \textbf{Sentimental Value}: Similarly, an important aspect of understanding a person's emotion is identifying the sentimental value that they assign to different memories and belongings.
    \begin{quote}
    Losing a T-shirt we wanted to throw out (low sentimental value) is unlikely to lead to \textit{sadness}, whereas it would be \textit{devastating} if the T-shirt was a gift from a lost family member (high sentimental value).
    \end{quote}
    \item \textbf{Persona}: we also wanted to analyze whether LLMs comprehend the reactions of people with pre-existing emotions. These could include phobias, appraisals (previous experiences), and personal traits (e.g., being antisocial or extroverted).
    \begin{quote}
    If a person with claustrophobia, who gets extremely uncomfortable in small or crowded spaces, is invited to a small space, they might experience \textit{fear}, but not when going to a spacious garden space.
    \end{quote}
    \begin{quote}
    \end{quote}
\end{itemize}

\subsection{Emotional Cues} 
Emotional intelligence enables us to recognize and understand cues about emotions of ourselves and others \cite{rivers2020emotional}. 
While recent research has shown that LLMs are capable of understanding and responding to direct and explicit emotional stimuli and cues \cite{li2023large}, it is not explored how such models would react to implicit cues.
To this end, we designed this category to assess LLM's comprehension of text-based vocal (e.g., vocal utterances, tone, and speech) and visual (e.g., facial/physical expressions) cues of emotions.
\begin{quote}
\textit{A person's face turning red could be a visual cue for being \textit{angry} or \textit{shy}.
A sigh could indicate \textit{relief} or \textit{annoyance}.}
\end{quote}

\subsection{Perspective Taking} 
Emotional understanding has significant correlations with affective theory-of-mind \cite{Mier2010Emo,kalbe2010dissociating, ferguson2010associations}, mainly in that they both require the ability to view situations from the perspective of others and simulate their emotions given the circumstances, formally known as perspective-taking.
Therefore, we adopt three of the prevalent tasks for assessing perspective-taking in theory-of-mind: \textit{False Belief}, \textit{Faux Pas}, \textit{Strange Story}.
However, contrary to the traditional implementation of these tests, our sole focus is on designing scenarios that trigger different emotions based on personal knowledge and views of the situation.
\begin{itemize}
    \item \textbf{Affective False Belief}: The Sally-Ann test \cite{baron1985does} is one of the de facto assessments for the theory of mind (ToM), i.e., the ability to infer the beliefs and mental states of others. 
    Recently, it has also been widely adopted for evaluating ToM in LLMs \cite{he2023hi, ma2023tomchallenges, kim-etal-2023-fantom} as it requires reasoning about each individual's knowledge and perspective on the situation to answer the corresponding questions.
    In our framework, we collected scenarios in which the individual's emotions could be implied through reasoning about their beliefs, which could be affected by trusting the word of others and/or being oblivious to certain events.
    \begin{quote}
    I was the only one who saw my friend's grades and realized that he failed the exam.
    Therefore, if I tell him that he passed the exam with flying colors, he would be \textit{excited}, not \textit{disappointed}.
    \end{quote}
    \item \textbf{Faux Pas}: Similarly, a more advanced assessment of ToM is conducted through the faux pas (i.e., tactless acts or remarks that cause unintentional negative consequences) detection test \cite{baron1999recognition}. 
    In this task, participants are presented with a social situation and are required to detect the presence and identify the faux pas.
    Inspired by this, we include a series of scenarios that include a faux pas and assess LLMs on identifying the emotions of the involved individuals. 
    In these scenarios, in addition to understanding social cues associated with a faux pas, LLMs also have to reason about each individual's beliefs and their known information to understand their emotions. 
    \begin{quote}
    If a person openly criticizes a painting without knowing it was drawn by their brother, then they may feel \textit{disgust} towards the painting and not \textit{embarrassment} due to their lack of information.
    \end{quote}
    \item \textbf{Strange Story}: Inspired by \citet{happe1994advanced}, we also designed scenarios that establish hypothetical grounds and imaginary assumptions that would contradict the normal pattern of behavior. 
    This further evaluates whether LLMs truly reason about the situation to infer the relevant emotions or base their judgments on learned patterns.
    \begin{quote}
    While getting an F in a test would regularly lead to \textit{disappointment}, getting an F in a class where the teacher only gives Fs to the highest mark leads to \textit{pride}.
    \end{quote}
\end{itemize}

\section{Experiment Prompts} \label{sec:prompts_appendix}
Our designed prompts are demonstrated in Table \ref{table:prompts}. 
For Chinese samples, we directly translated the provided prompts into Chinese.

\begin{table*}[!ht]
    \begin{tabular}{p{\linewidth}}
        \toprule
         \multicolumn{1}{c}{\textbf{System Prompt (Base)}} \\
        \midrule
        \textbf{**Instructions**} \\In this task, you are presented with a scenario, a question, and multiple choices. Please carefully analyze the scenario and take the perspective of the individual involved.\\\textbf{**Note**}\\Provide only one single correct answer to the question and respond only with the corresponding letter. \\Do not provide explanations for your response. \\
        \midrule
        \multicolumn{1}{c}{\textbf{System Prompt (CoT)}} \\ \midrule
        \textbf{**Instructions**}\\1. \textbf{**Reason**}: Read the scenario carefully, paying close attention to the emotions, intentions, and perspectives of the individuals involved. Then, using reason step by step by exploring each option's potential impact on the individual(s) in question. Consider their emotions, previous experiences mentioned in the scenario, and the possible outcomes of each choice.
        \\2. \textbf{**Conclude**} by selecting the option that best reflects the individual's perspective or emotional response. Your final response should be the letter of the option you predict they would choose, based on your reasoning. \\ \textbf{**Note**} \\ The last line of your reply should only contain the letter numbering of your final choice. \\
        \midrule
        \multicolumn{1}{c}{\textbf{Emotional Understanding (EU)}} \\ \midrule
        \textbf{For Emotions}\\
     Scenario: $[$scenario$]$  \\ Question: What emotion(s) would $[$subject$]$  ultimately feel in this situation? \\ Choices: $[$choices$]$  \\ 
        \textbf{For Causes}\\
        Scenario: $[$scenario$]$  \\ Question: Why would $[$subject$]$ feel $[$emotions$]$  in this situation? \\ Choices: $[$choices$]$  \\ 
        \midrule
        \multicolumn{1}{c}{\textbf{Emotional Application (EA)}} \\ \midrule
        Scenario: $[$scenario$]$  \\ Question: In this scenario, what is the most effective $[$problem type$]$ for $[$subject$]$? \\ Choices: $[$choices$]$  \\  \midrule
        \multicolumn{1}{c}{\textbf{Answer}} \\ \midrule
        \textbf{Without CoT} $\rightarrow$ Answer (Only reply with the corresponding letter numbering):\\
        \textbf{With CoT}$\rightarrow$Answer: Let's think step by step \\
\bottomrule
    \end{tabular}
  \caption{Our designed Prompts
  }
  \label{table:prompts}
\end{table*}

\section{Emotion Distribution}\label{sec:emo_dist}
Figure \ref{fig:topic_pic} demonstrates the category distribution for the collected samples.
\begin{figure}[!ht]
    \centering
    \includegraphics[width=\linewidth]{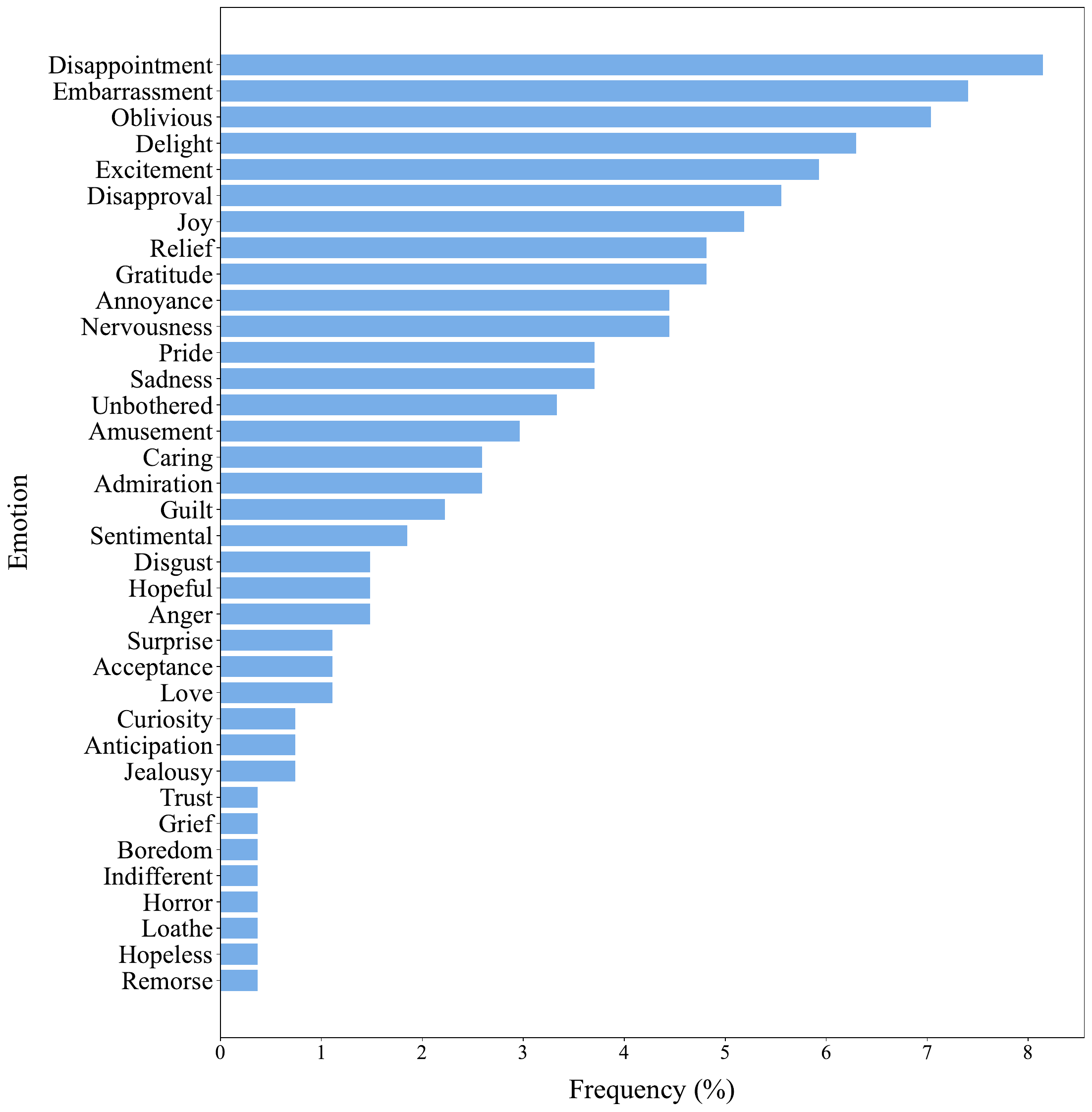}
    \caption{Emotion Distribution in \textsc{EmoBench}.}
    \label{fig:emotion_dist}
\end{figure}

\section{Human Evaluation} \label{sec:human_stats}
\begin{table}[!ht]
\resizebox{\linewidth}{!}{
    \begin{tabular}{c c c c}
        \toprule
        \multicolumn{2}{c}{} &
        \multicolumn{1}{c}{\textbf{EU}} & 
        \multicolumn{1}{c}{\textbf{EA}} 
        \\
        \multicolumn{2}{c}{} &
        \multicolumn{1}{c}{\textbf{($\mathbf{n=24}$)}}& 
        \multicolumn{1}{c}{\textbf{($\mathbf{n=24}$)}} 
        \\ \midrule
        \textbf{Gender, $\mathbf{n}$ (\%)} & M &  13 (54.17\%) & 8 (33.3\%) 
        \\ 
         & F & 11 (45.83\%)&  16 (66.67\%) \\
         \textbf{Age, \textit{Mean (SD)}}  &  & 23.42 (3.62)& 23.3 (1.98) \\
        \bottomrule
    \end{tabular}
    }
    \caption{
    Demographics of Our Human Participants ($n=48$). M and F indicate Male and Female, respectively.
  }
  \label{table:user_demo}
\end{table}

During registration for our experiments, all candidates disclosed their demographics, language, and task preferences.
As a part of our annotation quality control, we excluded individuals under the age of 21 as a means of ensuring emotional maturity (the ability to understand and manage emotions; \citet{jobson2020emotional}).
In addition, we required each candidate to correctly answer all of the questions (six MCQs) in a randomly sampled subset of our benchmark.

A total of 70 individuals registered for our experiment. 
From this candidate pool, we recruited 48 participants (31.43\% rejection rate) based on the above criteria and their pre-disclosed language-task preferences.
Our participants' demographics are summarized in Table \ref{table:user_demo}.
All the candidates were informed of the purpose of our study and consented to participate in our experiments.
Accordingly, we allocated an equal number of candidates to each language-task evaluation pair ($n=12$).
Each participant was compensated 14.28\$ per hour, which is well over the minimum wage in the US\footnote{\url{www.dol.gov/general/topic/wages/minimumwage}}.
Our guidelines are provided in Figures \ref{eu_guideline} and \ref{ea_guideline}.

\onecolumn
\phantomsection
\label{fig:EU_guidelines}
\begin{tcolorbox}[colback=white!95!gray,colframe=gray!50!black,rounded corners,label={eu_guideline}, title={Emotional Understanding Guideline}]
\begin{lstlisting}[breaklines=true, xleftmargin=0pt, breakindent=0pt, columns=fullflexible, mathescape=true, basicstyle=\large]
Background
\end{lstlisting}
\begin{lstlisting}[breaklines=true, xleftmargin=0pt, breakindent=0pt, columns=fullflexible, mathescape=true]
In this test, 
(1) You will be presented with 30 emotional scenarios.
(2) You will be asked to identify the emotions of the individual and their causes in this scenario.
Your task is to :
(1) Carefully read the design section and familiarize yourself with the emotion category. 
(2) Take the perspective of the people involved (think how you would feel in this situation).
(3) Choose the appropriate answer from the given choices and enter in the provided Excel sheet.
\end{lstlisting}
\begin{lstlisting}[breaklines=true, xleftmargin=0pt, breakindent=0pt, columns=fullflexible, mathescape=true, basicstyle=\large]
Emotion Taxonomy
\end{lstlisting}
\begin{lstlisting}[breaklines=true, xleftmargin=0pt, breakindent=0pt, columns=fullflexible, mathescape=true]
[BASIC EMOTIONS]: Our emotion category includes 8 basic emotions: Sadness, Anger, Joy, Fear, Anticipation, Trust, Disgust, and Surprise.
[MIXED EMOTIONS]: By combining the above basic emotions, we can get 14 mixed emotions: Guilt (joy + fear), Pride (joy + anger), Excitement/Hopeful (Optimism) (joy + anticipation), Love/Caring/Gratitude (joy + + trust), Amusement/Delight (joy + surprise), Disapproval/Disappointment (surprise + sadness), Sentimental(trust + sadness), Jealousy (sadness + anger), Pessimism (anticipation + sadness), Remorse (disgust + sadness), Hopeless (fear + sadness), Embarrassment (fear + disgust), Nervousness (fear + anticipation), Curiosity (trust + surprise).
[NEUTRAL]: In case the individual in the situation is not experiencing any emotions, we would label them as 1) unbothered (indifferent) or 2) Oblivious, depending on the situation.


\end{lstlisting}
\begin{lstlisting}[breaklines=true, xleftmargin=0pt, breakindent=0pt, columns=fullflexible, mathescape=true, basicstyle=\large]
Example
\end{lstlisting}
\begin{lstlisting}[breaklines=true, xleftmargin=0pt, breakindent=0pt, columns=fullflexible, mathescape=true]
Scenario: James and I are coworkers. We've been best friends for over a decade. Our boss gives out an employee of the year award every year. This year, we both applied as candidates for this reward and worked hard to get it. The results were announced yesterday. James won the award. 
Question 1: Ultimately, what are the emotions that I would feel in this scenario?
Choices:
A) Disappointment & Remorse      B) Pride & Remorse     C) Disappointment & Indifferent
D) Disappointment & Admiration      E) Amusement & Indifferent      F) Admiration & Pride

Question 2: Why would I feel these emotions in this scenario?
A) I am upset that my friend won the award instead of me & I am convinced that our boss was biased in his decision
B) I am convinced that our boss was biased in his decision & I care for James as my best friend and believe he worked hard to win the award
C) I think I wasn't good enough to win the award & I am convinced that our boss was biased in his decision
D) I am upset that my friend won the award instead of me & I admire our boss for making an unbiased decision
E) I think I wasn't good enough to win the award & I care for James as my best friend and believe he worked hard to win the award
F) I am upset that my friend won the award instead of me & I care for James as my best friend and believe he worked hard to win the award

Answer: D & E

\end{lstlisting}

\end{tcolorbox}

\onecolumn
\phantomsection
\label{fig:EA_guidelines}
\begin{tcolorbox}[colback=white!95!gray,colframe=gray!50!black,rounded corners,label={ea_guideline}, title={Emotional Application Guideline}]
\begin{lstlisting}[breaklines=true, xleftmargin=0pt, breakindent=0pt, columns=fullflexible, mathescape=true, basicstyle=\large]
Background
\end{lstlisting}
\begin{lstlisting}[breaklines=true, xleftmargin=0pt, breakindent=0pt, columns=fullflexible, mathescape=true]
In this test, 
(1)You will be presented with 30 emotional scenarios 
(2)You will be asked to identify the most effective action/response in this scenario
Instruction
Your task is to :
(1) Carefully read the presented scenarios.
(2) Take the perspective of the people involved in the scenario to understand what you would do in this situation.
(3) Now, think what you should do after understanding and managing your emotions.
(4) Choose the appropriate answer from the given choices.
(5) Enter the chosen answer in the provided Excel sheet
(6) Rename the file to "{name}.xlsx", where {name} is replaced with your name.
(7) Submit your answers to [link]
\end{lstlisting}
\begin{lstlisting}[breaklines=true, xleftmargin=0pt, breakindent=0pt, columns=fullflexible, mathescape=true, basicstyle=\large]
Example
\end{lstlisting}
\begin{lstlisting}[breaklines=true, xleftmargin=0pt, breakindent=0pt, columns=fullflexible, mathescape=true]
Scenario: Robert had an old red t-shirt he wanted to throw out. His friend Andrew asked to borrow the T-shirt for a party. The next day, Andrew came to Robert and told him that he lost it.
Question: What is the most effective action for Robert in this scenario?
Choices:
A) Express forgiveness and understanding
B) Request a replacement of a similar value or style
C) Mention that it's okay as the t-shirt didn't have any value to him
D) Choose not to lend anything to Andrew in the future

Answer: C

Note: In cases where multiple options are plausible, choose the most likely/useful one
\end{lstlisting}

\end{tcolorbox}

\end{document}